%% file: template.tex
\RequirePackage{fix-cm}
\documentclass[smallextended]{svjour3}       
\smartqed  
\usepackage{graphicx}
\usepackage{tabularx}
\usepackage[numbers]{natbib}
\usepackage{color}
\usepackage{multirow}
\usepackage{subfig} 
\usepackage{ctable}
\usepackage{subfiles}
\usepackage{blindtext}
\usepackage[]{algorithm2e}

\begin{document}

\title{Effective deep learning training for  single-image super-resolution in endomicroscopy exploiting video-registration-based reconstruction}


\author{ Daniele Rav\`{\i}$^\dagger$ \and
       Agnieszka Barbara Szczotka$^\dagger$\thanks{$^\dagger$ DR and ABS contributed equally to this work}\and
        Dzhoshkun Ismail Shakir \and
        Stephen P Pereira \and
        Tom Vercauteren
}
\authorrunning{Daniele Rav\`{\i}, Agnieszka Barbara Szczotka et al.}
\titlerunning{Effective deep learning training for  single-image super-resolution}

%

\date{Received: date / Accepted: date}

\institute{
D. Rav\`{\i}, A. B. Szczotka, D. I. Shakir, T. Vercauteren \at
Wellcome / EPSRC Centre for Interventional and Surgical Sciences, University College London
\email{d.ravi@ucl.ac.uk, agnieszka.szczotka.15@ucl.ac.uk}
\and
S. P Pereira  \at
UCL Institute for Liver and Digestive Health
}

\maketitle

\begin{abstract}
\subfile{abstract}

\keywords{Example-Based Super-resolution \and Deep Learning  \and Probe-based  Confocal  Laser  Endomicroscopy \and Mosaicking}
\end{abstract}

\section{Introduction}
\subfile{introduction}
\section{Related work}
\subfile{related}
\section{Materials and methods}
\subfile{methods}
\section{Results}
\subfile{results}
\section{Discussion and conclusions}
\subfile{disscusion}


\section*{Acknowledgement}

\textbf{Funding:} This work was supported by Wellcome/EPSRC [203145Z/16/Z; NS/A000050/1; WT101957; NS/A000027/1; EP/N027078/1]. This work was undertaken at UCL and UCLH, which receive a proportion of funding from the DoH NIHR UCLH BRC funding scheme.
\textbf{Conflict of Interest:} The PhD studentship of Agnieszka Barbara Szczotka is funded by Mauna Kea Technologies, Paris, France. Tom Vercauteren owns stock in Mauna Kea Technologies, Paris, France.
The other authors declare no conflict of interest.
\textbf{Ethical approval:}  All procedures performed in
studies involving human participants were in accordance with the ethical standards of the
institutional and/or national research committee and with the 1964 Helsinki declaration and its
later amendments or comparable ethical standards.
\textbf{Informed consent:} For this type of study formal consent is not required. This article does not contain patient data.
\textbf{Thanks:}
The authors would like to thank the High Dimensional Neurology group, Institute of Neurology, UCL for provide computational support. The authors would like to thank the independent experts at Mauna Kea Technologies for participating to the MOS survey.


\bibliographystyle{spbasic}      
\bibliography{template}   

%
%

\end{document}

%% file: abstract.tex
\label{abstract}
\textbf{Purpose:} Probe-based Confocal Laser Endomicroscopy (pCLE) is a recent imaging modality that allows performing in vivo optical biopsies. The design of pCLE hardware, and its reliance on an optical fibre bundle, fundamentally limits the image quality with a few tens of thousands fibres, each acting as the equivalent of a single-pixel detector, assembled into a single fibre bundle. Video-registration techniques can be used to estimate high-resolution (HR) images by exploiting the temporal information contained in a sequence of low-resolution (LR) images.
However, the alignment of LR frames, required for the fusion, is computationally demanding and prone to artefacts.
\textbf{Methods:}
In this work, we propose a novel synthetic data generation approach to train exemplar-based Deep Neural Networks (DNNs). HR pCLE images with enhanced quality are recovered by the models trained on pairs of estimated HR images (generated by the video-registration algorithm) and realistic synthetic LR images.
Performance of three different state-of-the-art DNNs techniques were analysed on a Smart Atlas database of 8806 images from 238 pCLE video sequences. The results were validated through an extensive Image Quality Assessment (IQA) that takes into account different quality scores, including a Mean Opinion Score (MOS). 
 \textbf{Results:}
Results indicate that the proposed solution produces an effective improvement in the quality of the obtained reconstructed image.
 \textbf{Conclusion:}
The proposed training strategy and associated DNNs allows us to perform convincing super-resolution of pCLE images.

%% file: introduction.tex
\label{intro}
Probe-based confocal laser endomicroscopy (pCLE) is a state-of-the-art imaging system used in clinical practice for in situ and real-time in vivo optical biopsy.
In particular, recent works using Cellvizio (Mauna Kea Technologies, France) have demonstrated the impact of introducing pCLE as a new imaging modality for the diagnostics procedures of conditions such as pancreatic cystic tumours and the surveillance of Barrett's oesophagus~\cite{Fugazza2016}. pCLE is a recent imaging modality in gastrointestinal and pancreaticobiliary diseases~\cite{Fugazza2016}.
\par 
The authors of~\cite{Fugazza2016} have shown that despite clear clinical benefits of pCLE, improving its specificity and sensitivity would help it become a routine diagnostic tool. 
Specificity and sensitivity are directly dependent on the quality of the pCLE images. Therefore, increasing the resolution of these images might bring a more reliable source of information and improve pCLE diagnosis.
\par

Certainly, the key point of pCLE is its suitability for real-time and intraoperative usage. Having high-quality images in real-time potentially allows
for better pCLE interpretability. Thus, offline processing would not fit in the standard clinical work-flow required in this context.
\par
The trend for image sensor manufacturers is to increase the resolution, as apparent in the current move to high-definition endoscopic detectors. Recently introduced 4K endoscopes provide 8M pixels, a difference to pCLE of 2-to-3 orders of magnitude. In pCLE, reliance on an imaging guide - an optical fibre bundle, composed of a few tens of thousands of optical fibres, each acting as the equivalent of a single-pixel detector - fundamentally limits the image quality. These fibres are irregularly positioned in the bundle which implies that tissue signal is a collection of pixels sampled on an irregular grid. Hence, a reconstruction procedure is needed for mapping the irregular samples to a Cartesian image.
Other factors that reduce pCLE image quality are cross-talk among neighbouring fibres and limited signal to noise ratio. All these factors lead to the generation of images with artefacts, noise, relatively low contrast and resolution.
This work proposes a software-based resolution augmentation method which is more agile and simpler to implement than hardware engineering solutions.
\par
Building on from the idea that high-resolution (HR) images are desired, this study explores advanced
single-image super-resolution (SISR) techniques which can contribute to effective improvement in image quality. Although SISR for natural images is a relatively mature field, this work is the first attempt to translate these solutions into the pCLE context. Beyond SISR, video-registration technique~\cite{vercauteren2006robust} have been proposed to increase the resolution of pCLE. Such methods provide a baseline super-resolution technique, but suffers from artefact and are computationally too expensive to be applied in real-time. Because of the recent success of deep learning for SISR on natural images~\cite{timofte2017ntire},
this work focuses on exemplar-based super-resolution (EBSR) deep learning techniques. However, the translation of these methods to the pCLE domain is not straightforward, notably due to the lack of ground truth HR images required for the training. There is indeed no equivalent imaging device capable of producing higher resolution endomicroscopic imaging, nor any robust and highly accurate means of spatially matching microscopic images acquired across scales with different devices. Furthermore, in comparison to natural images,  currently available pCLE images suffer from specific artefacts introduced by the reconstruction procedure that maps the tissue signal from the irregular fibre grid to the Cartesian grid.
\par
The contribution of this work is three-fold.
First, three deep learning models for SISR are examined on the pCLE data. Second, to overcome the problem of the lack of ground truth low-resolution (LR)/HR image pairs for training purposes,
a novel pipeline to generate pseudo-ground-truth data by leveraging an existing video-registration technique \cite{vercauteren2006robust} is proposed.

Third, in the absence of a reference HR ground truth, to assess the clinical validity of our approach, a Mean Opinion Score (MOS) study was conducted with nine experts (1-10 years of experience) each assessing 46 images according to 3 different criteria. To our knowledge, 
this is the first research work to: address the challenge of SISR reconstruction for pCLE images based on deep learning; generate pCLE pseudo-ground-truth data for training of EBSR models and; demonstrate that pseudo-ground-truth trained models provide convincing SR reconstruction.
\par
The rest of the paper is organised as follows.
Section~\ref{Related} presents the state of the art for SISR with natural images. Section~\ref{Materials and methods} presents the proposed training methodology based on realistic pseudo-ground-truth generation and detail the implementation of the SISR models. 
Section~\ref{results} gives information on the evaluation of our approach using a quantitative image quality assessment (IQA) and a MOS study.
Section \ref{disscusion} summarises the contribution of this research  to pCLE SISR.

%% file: related.tex
\label{Related}
Super-resolution (SR) has received a lot of interest from the computer vision community in the recent decades~\cite{park2003super}. Initial SR approaches were based on single-image super-resolution (SISR) and exploited signal processing techniques applied to the input image.
An alternative to SISR is multi-frame image super-resolution based on the idea that HR image can be reconstructed by fusing many LR images together. Ideally, the combination of several LR image sources enriches the information content of the reconstructed HR image and contributes to improving its quality. Registration can be used to merge LR images acquired at slightly shifted field-of-views into a unified HR image.\par
In the specific context of pCLE, the work proposed by Vercauteren et al.~\cite{vercauteren2006robust} presents a video-registration algorithm that, in some cases, can improve spatial information of the reconstructed pCLE image, and reveals details which were not visible initially. The quality of the registration is a key step to the success of the SR reconstruction, but the alignment of images captured at different times is not trivial. Misalignment leads to incorrect fusion and generates artefacts such as ghosting.
Moreover, registration is a computationally expensive technique, making this approach unsuitable for real-time purposes.
\par
Another interesting approach to SISR is exemplar-based super-resolution (EBSR), which learns the correspondence between low- and the high-resolution images. Thanks to the recent success of deep learning and Convolutional Neural Networks (CNNs), EBSR methods currently represent the state-of-the-art for the SR task~\cite{timofte2017ntire}. 
Although many research groups have worked on deep-learning-based SR for natural images, and although CNNs are currently widely used in various medical imaging problems~\cite{ravi2017deep}, only recently have CNNs been used for SR in medical imaging. Noteworthy is the work proposed in ~\cite{tanno2017bayesian} that attempt to improve the quality of magnetic resonance images.
 \par
The behaviour of CNNs, especially in the context of SR, is strongly driven by the choice of a loss function, and the most popular one is mean squared error (MSE)~\cite{zhao2015loss}. Although MSE as a loss function steers the SR models towards the reconstitution of HR images with high peak signal-to-noise ratios, this does not necessarily mean that the final images will provide a good quality perception. A model trained with a selective loss function involving a Generative Adversarial Network for Image Super-Resolution (SRGAN) was proposed by Ledig et al.~\cite{ledig2016photo}. The authors designed an adversarial loss to classify HR images into SR images and ground-truth HR images.
Based on a MOS study, the authors showed that the participants perceived the quality of the restored HR images as higher compared to the image quality measured only by a PSNR.  
\par
Another critical issue with deep CNNs is the convergence speed. Several solutions, such as using a very high learning rate for network training~\cite{kim2016accurate}, and removing batch-normalisation modules~\cite{lim2017enhanced} were proposed to tackle this issue.

%% file: methods.tex
\label{Materials and methods}
The Smart Atlas database~\cite{andre2011smart}, a collection of 238 anonymised pCLE video sequences of the colon and oesophagus, is used in this study.
The database was split into three subsets: training set (70\%), validation set (15\%), and test set (15\%). Each subset was created ensuring that colon and oesophagus tissue were equally represented. Data were acquired with 23 unique probes of the same pCLE probe type. The SR models are specific to the type of the probe but generic to the exact probe being used. Thus, the models do not need to be retrained for probes of the same type. Another type of probe, such as needle-CLE (nCLE) would require a specifically trained model. nCLE and pCLE differ by the number of optical fibres and the design of the distal optics.

Section~\ref{registration} explains
how the pseudo-ground-truth HR images were generated. Section~\ref{syntetic} describes our proposed simulation framework to generate synthetic LR 
($LR_{syn}$) images from original LR ($LR_{org}$) images.

Section~\ref{procesing} presents the pre-processing steps needed for standardising the input images and details the implementation of the super-resolution CNNs used in this study.
\subsection{Pseudo-ground-truth image estimation based on video registration}
\label{registration}
To compensate for the lack of ground-truth HR pCLE data, a registration-based mosaicking technique~\cite{vercauteren2006robust} was used to estimate HR images. 
Mosaicking acts as a classical SR technique and fuses several registered input frames by averaging the temporal information. The mosaics were generated for the entire Smart Atlas database and used as a source of HR frames.
\par
Since mosaicking generates a single large field-of-view mosaic image from a collection of input LR images, it does not directly provide a matched HR image for each LR input. To circumvent this, we used the mosaic-to-image diffeomorphic spatial transformation resulting from the mosaicking process to propagate and crop the fused information from the mosaic back into each input LR image space. The image sequences resulting from this method are regarded as estimates of HR frames. These estimates will be referred to as $\widehat{HR}$ in the text.
\par
The image quality of the mosaic image heavily depends on the accuracy of the underpinning registration which is a difficult task.
The corresponding pairs of LR and $\widehat{HR}$ images generated by the proposed registration-based method suffer from artefacts, which can hinder the training of the EBSR models.
\begin{figure*}
\begin{center}
  \includegraphics[width=0.90\textwidth,trim={0 7cm 0 1.2cm},clip]{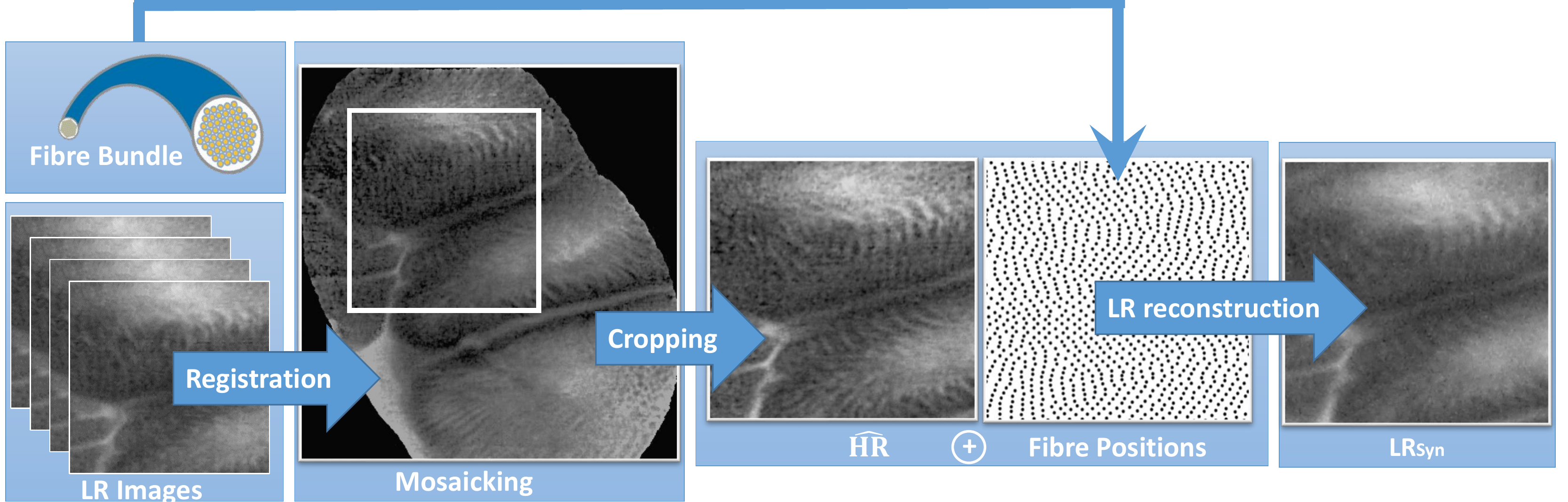}
\caption{Pipeline used to generate LR synthetic images}
\label{fig:Mosaic}       
\end{center}
\end{figure*}
\par
Specifically, it can be observed that alignment inaccuracies occurring during mosaicking were a source of ghosting artefacts and that in combination with residual misalignments between the LR and $\widehat{HR}$ images, creates unsuitable data for the training.
Sequences with obvious artefacts were manually discarded. However, even on this selected data-set, training issues were observed. To address these, we simulated LR-HR image pairs for training EBSR algorithms while leveraging the registration-based $\widehat{HR}$ images as realistic HR images. \par

\subsection{Generation of realistic synthetic pCLE data}
\label{syntetic}
Currently available pCLE images are reconstructed from scattered fibre signal.
Every fibre in the bundle acts as a single pixel detector.
To reconstruct pCLE images on a Cartesian grid, Delaunay triangulation and piecewise linear interpolation are used. The simulation framework developed in this study mimicks the standard pCLE reconstruction algorithm and starts by assigning to each fibre the average of the signal from seven neighbouring pixels~\cite{le2004towards}. In the standard reconstruction algorithm, the fibre signal, which includes noise, is then interpolated. Similarly, noise was added to the simulated data to produce realistic images and avoid creating a wide domain gap between real and simulated pCLE images. 

\par
Despite some misalignment artefacts, the registration-based generation of $\widehat{HR}$ presented in Section~\ref{registration} produces images with fine details and a high signal-to-noise ratio. Our simulation framework uses these $\widehat{HR}$ and produces simulated LR images with a perfect alignment.
\par
The proposed simulation framework relies on observed irregular fibre arrangements and corresponding Voronoi diagrams. Each fibre signal was extracted from an $\widehat{HR}$ image, by averaging the $\widehat{HR}$ pixel values within the corresponding Voronoi cell.
\par

To replicate realistic noise patterns on the simulated LR images, additive and multiplicative Gaussian noise ($a$ and $m$ respectively) is added to the extracted fibre signals $fs$ to obtain a noisy fibre signal $nfs$ as:
$nfs =(1+m).*fs + a $.
The standard deviation of the noise distributions were tuned based on visual similarity between $LR_{org}$ and $LR_{syn}$ and between their histograms.
Sigma values were 0.05 and 0.01*$(max \ fs -min \ fs)$ for multiplicative and additive Gaussian distribution respectively.

In the last step, Delaunay-based linear interpolation was performed thereby leading to our final simulated LR images.

\par
LR and $\widehat{HR}$ images were combined into two data-sets: 1. Original pCLE ($pCLE_{org}$) built with pairs of $LR_{org}$ taken from sequences of Smart Atlas database and $\widehat{HR}$ images, and 2. synthetic pCLE ($pCLE_{syn}$) built by replacing the $LR_{org}$ images with $LR_{syn}$ images.

\subsection{Implementation details}
\label{procesing}

The data-sets were pre-processed in three steps. First, intensity values were normalised: $LR = LR - mean_{lr}/std_{lr}$ and $HR = HR - mean_{lr}/std_{lr}$. Second, pixels values were scaled of every frame individually in the range [0-1].
Third, non-overlapping patches of $64 \times 64$ pixels were extracted for the training phase,
considering only pixels in the pCLE Field of View (FoV). A stochastic patch-based training was used for training the networks, with a minibatch of size 54 
patches to fit into the GPU memory (12GB).
\par

Models were trained with patches from the training set. The patches from the validation set were used to monitor the loss during training with the purpose to avoid overfitting.
Since all the considered networks are fully convolutional, the test images were processed full-size
and no patch processing is required during the inference phase.

\par
Three CNNs networks for SR were used: sparse-coding based FSRCNN~\cite{dong2016accelerating}, residual based EDSR~\cite{lim2017enhanced}, and generative adversarial network SRGAN~\cite{ledig2016photo}. Every model was trained with the two datasets presented in section~\ref{syntetic}.
\par
MSE is the most commonly used loss function for SR. Zhao et al. ~\cite{zhao2015loss} showed that MSE has two limitations: it does not converge to the global minimum and produces blocky artefacts. In addition to demonstrating that L1 loss outperforms L2, the authors also introduced a new loss function SSIM+L1 by incorporating the Structural Similarity
(SSIM)~\cite{wang2004image}.
FSRCNN and EDSR were trained considering independently both L1 and SSIM+L1 to investigate their applicability for our data based on a quantitative comparison. 

%% file: results.tex
\label{results}
Acknowledging the lack of proper ground truth for super-resolution of pCLE and the ambiguous nature of established IQA metrics, a three-stage approach was designed for the evaluation of the proposed method using the three SR architectures considered in Section~\ref{Materials and methods}. \par
The first stage, presented in Section~\ref{synteticResults} and relying on the quantitative assessment, demonstrates the applicability of EBSR for pCLE in the ideal synthetic case where ground-truth is available. In this quantitative stage, the inadequacy of the existing video-registration-based high-resolution images as a ground truth for EBSR training purpose is demonstrated. \par
The second stage, presented in Section~\ref{RealDataResults} focuses on the quantitative assessment of our methods in the context of real input images and on the evaluation of our best model against other state-of-the-art SISR methods. \par
In the third stage, performed to overcome the limitations of the quantitative assessment, a MOS study was carried out by recruiting nine independent experts, having 1-10 years of experience working with pCLE images.

\begin{table*}
\caption{Quantitative results obtained on full-size images from the test set for different training and testing strategies}.
\label{table:trainResults}
\begin{center}
\resizebox{\textwidth}{!}{%

\begin{tabular}{c|c|c|c>{\centering\arraybackslash}p{1.5cm}>{\centering\arraybackslash}p{1.5cm}>{\centering\arraybackslash}p{1.5cm}>{\centering\arraybackslash}p{1.6cm}>{\centering\arraybackslash}p{1.5cm}c}

   &Train&Test& LR  &    EDSR L1&    FSRCNN L1&    EDSR SSIM+L1&    FSRCNN SSIM+L1&    SRGAN \\
\hline
SSIM with $\widehat{HR}$ &\multirow{4}{*}{\rotatebox{65}{$pCLE_{syn}$}}&\multirow{4}{*}{\rotatebox{65}{$LR_{syn}$}}&    0.81$\pm$0.06&        \textbf{0.87$\pm$0.06}&    0.86$\pm$0.06&    \textbf{0.87$\pm$0.06}&    0.86$\pm$0.06&    0.76$\pm$0.06\\
$\Delta$ GCF with $\widehat{HR}$ &&&    -0.42$\pm$0.31&        -0.22$\pm$0.13&    -0.26$\pm$0.16&    -0.1$\pm$0.13&    \textbf{-0.06$\pm$0.16}&    -0.09$\pm$0.34\\
$\Delta$ GCF with LR &&&    0$\pm$0&        0.21$\pm$0.31&    0.17$\pm$0.19&    0.32$\pm$0.32&    \textbf{0.36$\pm$0.23}&    0.34$\pm$0.22\\
\cline{4-9}
 $Tot_{cs}$&&& 0.46&	0.67&	0.63&	\textbf{0.71}&	\textbf{0.71}&	0.47\\
 
\specialrule{.2em}{.1em}{.1em} 

 SSIM with $\widehat{HR}$ &\multirow{4}{*}{\rotatebox{65}{$pCLE_{org}$}}&\multirow{4}{*}{\rotatebox{65}{$LR_{org}$}}&    0.81$\pm$0.06&    \textbf{0.83$\pm$0.06}&    0.82$\pm$0.06&    0.82$\pm$0.06&    0.82$\pm$0.06&    0.75$\pm$0.05\\
$\Delta$ GCF with $\widehat{HR}$&&&    -0.24$\pm$0.37&        -0.24$\pm$0.29&    -0.15$\pm$0.3&    -0.11$\pm$0.3&    \textbf{-0.01$\pm$0.32}&    -0.1$\pm$0.37\\
$\Delta$ GCF with LR&&&    0$\pm$0&    0.01$\pm$0.13&    0.09$\pm$0.11&    0.13$\pm$0.13&    \textbf{0.24$\pm$0.12}&    0.14$\pm$0.15\\
\cline{4-9}
$Tot_{cs}$&&& 0.44&	0.50&	0.52&	0.54&	\textbf{0.57}&	0.37
 \\
\specialrule{.2em}{.1em}{.1em}

SSIM with $\widehat{HR}$&\multirow{4}{*}{\rotatebox{65}{$pCLE_{syn}$}}&\multirow{4}{*}{\rotatebox{65}{$LR_{org}$}}&    0.81$\pm$0.06&        0.81$\pm$0.06&    \textbf{0.82$\pm$0.06}&    0.80$\pm$0.06&    0.81$\pm$0.06&    0.75$\pm$0.06\\
$\Delta$ GCF with $\widehat{HR}$&&&    -0.24$\pm$0.37&        0.18$\pm$0.29&    0.05$\pm$0.27&    \textbf{0.33$\pm$0.31}&    0.23$\pm$0.28&    -0.04$\pm$0.44\\
$\Delta$ GCF with LR&&&    0$\pm$0&    0.42$\pm$0.35&    0.29$\pm$0.22&    \textbf{0.57$\pm$0.36}&    0.47$\pm$0.26&    0.21$\pm$0.26\\
\cline{4-9}
$Tot_{cs}$&&& 0.44 &	0.61&	0.58&	\textbf{0.65}&	0.64&	0.38
\\

\end{tabular}
}
\end{center}
\end{table*}

\subsection{Quantitative analysis}
\label{QuantitativeAnalisi}
For the quantitative analysis, the SR images were examined exploiting two complementary metrics: i) SSIM to evaluate the similarity between the SR image and the $\widehat{HR}$, and ii) Global Contrast Factor (GCF)~\cite{matkovic2005global} as a reference-free metric for measuring image contrast which is one of the key characteristic of image quality in our context. Analysing both SSIM and GCF in combination leads to a more robust evaluation. SSIM alone cannot be depended on when the reference image is unreliable while improvements in GCF alone can be achieved deceitfully for example by adding a large amount of noise.
\par
Using these metrics, {six scores for each SR method were extracted}: mean and standard deviation of i) SSIM between SR and $\widehat{HR}$, ii) GCF differences between SR and LR and iii) GCF differences between SR and the $\widehat{HR}$.
Finally, to determine which approach performs better, a composite score $Tot_{cs}$ obtained by averaging the normalised value of SSIM with the normalised GCF difference between SR and LR was defined. Both factors are re-scaled to the range [0,1].
In our quantitative assessment, the score obtained by the initial $LR_{org}$ was considered as baseline reference.

\subsection{Experiments on synthetic data}
\label{synteticResults} 
In the first experiment, synthetic data are used to demonstrate that our models work in the ideal situation where ground truth is available. The first section of Table~\ref{table:trainResults} shows the scores obtained when the SR models are trained on $pCLE_{syn}$ and tested on $LR_{syn}$. Here it is evident that the EDSR and FSRCNN trained with SSIM+L1 obtain a substantial improvement on the different quality factors with respect to the LR image. More specifically, in comparison with the initial LR image, the SSIM was increased by +0.06 when EDSR is used and by +0.05 when FSRCNN is used. These approaches also yield a GCF value that is very close to the GCF in $\widehat{HR}$ and an improvement of +0.32 and +0.36 in the GCF with respect to LR images. Statistical significance of these improvements was assessed with a paired t-test (p-value less than 0.0001). From this experiment, it is possible to conclude that the proposed solution is capable of performing SR reconstruction when the models are trained on synthetic data  with no domain gap at test time.

\begin{figure*}
  \begin{center}
      \includegraphics[width=0.5\textwidth,trim={4cm 1cm 10.5cm 0.5cm},clip]{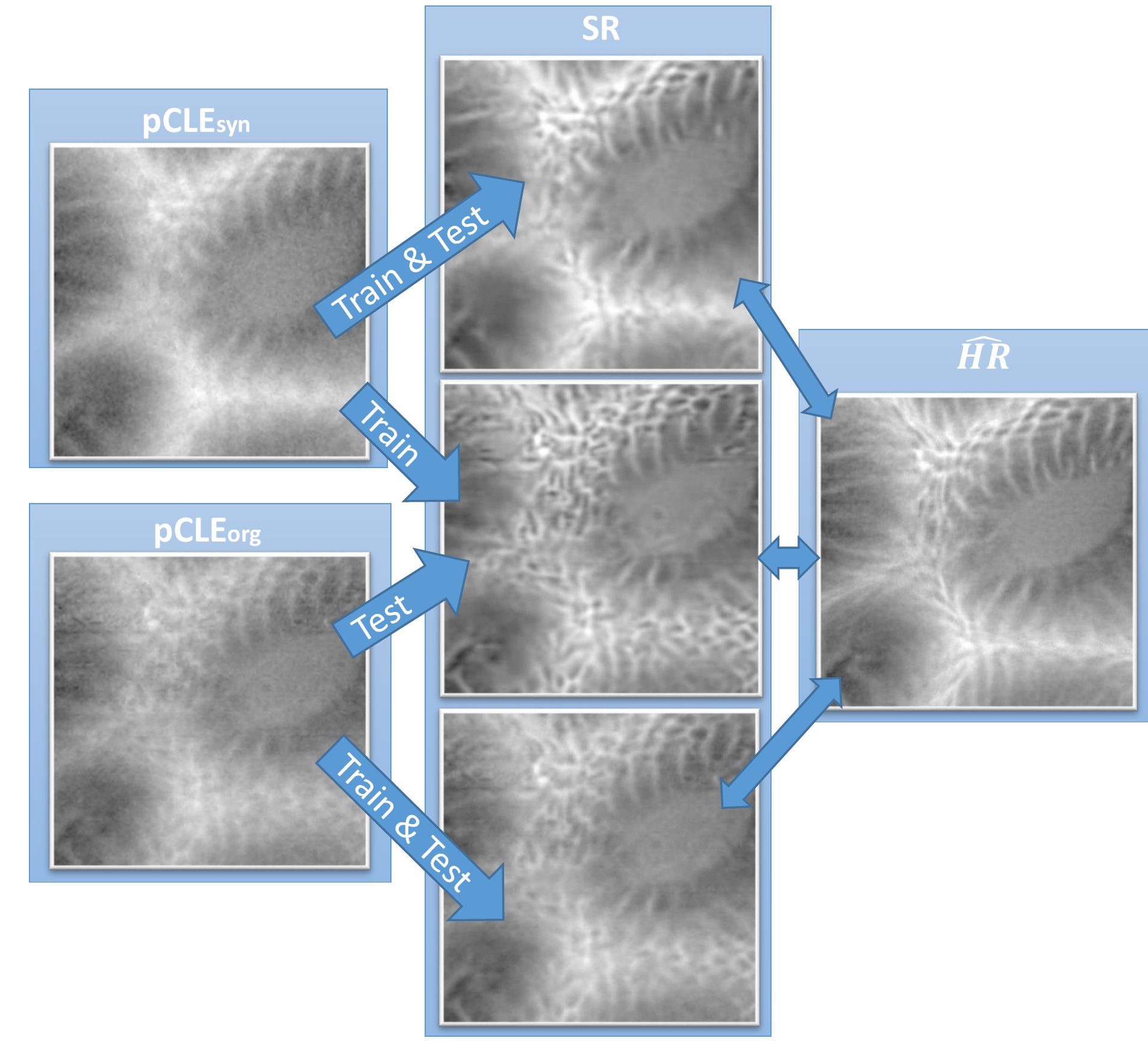}

    \caption{Example of SR images obtained when $pCLE_{syn}$ and $pCLE_{org}$ are used for train and test. From top to the bottom, the images in the middle represent the SR image obtained when: i) $pCLE_{syn}$ are used for train and test, ii) $pCLE_{syn}$ are used for train, and the $pCLE_{org}$ are used for test, and iii) $pCLE_{org}$ are used for train and test.}
    \label{fig:diffentTraining}     
\end{center}
\end{figure*}

\subsection{Experiments on original data}
\label{RealDataResults} 

When real images are considered, the same conclusions cannot be reached. The results obtained by training on $pCLE_{org}$ and testing on $LR_{org}$ are reported in the second section of Table~\ref{table:trainResults} and here it is evident that all the different quality factors decrease. The best approach is the FSRCNN trained using SSIM+L1 as loss function. With respect to the previous case this approach loses 0.04 on the SSIM, and 0.12 on the $\Delta$ GCF with LR. This leads to a final reduction of 0.14 for the $Tot_{cs}$ score. In this scenario, the deterioration of SSIM and GCF compared to the previous synthetic case can be due to the use of inadequate
$\widehat{HR}$ images during the training
(i.e. misalignment during the fusion, lack of compensation for motion deformations, etc.). Better results are instead obtained when the SR models performed on $LR_{org}$ images are trained using the $pCLE_{syn}$ (last section of Table~\ref{table:trainResults}).
Here, the quality factors increased when compared to the previous case, although they do not overcome the results obtained when the approach is trained and tested on synthetic data. EDSR, in particular, has a $Tot_{cs}$ score of 0.65 that is 0.08 better than the best approach trained on $pCLE_{org}$ (the second section of Table~\ref{table:trainResults}) and 0.06 worse than the best approach trained and tested on $pCLE_{syn}$ (first section of Table~\ref{table:trainResults}). The GCF obtained here are in general much better when compared to the previous two cases. An example of the visual results from the different training modalities is shown in Fig.~\ref{fig:diffentTraining}.
In conclusion, our findings suggest that existing video-registration-based approaches are inadequate to serve as a ground truth for HR images, while EBSR approaches, such as the EDSR and FSRCNN when trained on synthetic data can produce SR images that enhance the quality of the LR images.
\par
Due to our conclusions, the MOS study was performed using images obtained
from the models trained only with synthetic data.

\begin{figure*}
  \begin{center}

    \includegraphics[width=0.6\textwidth,trim={0 0 14cm 0},clip]{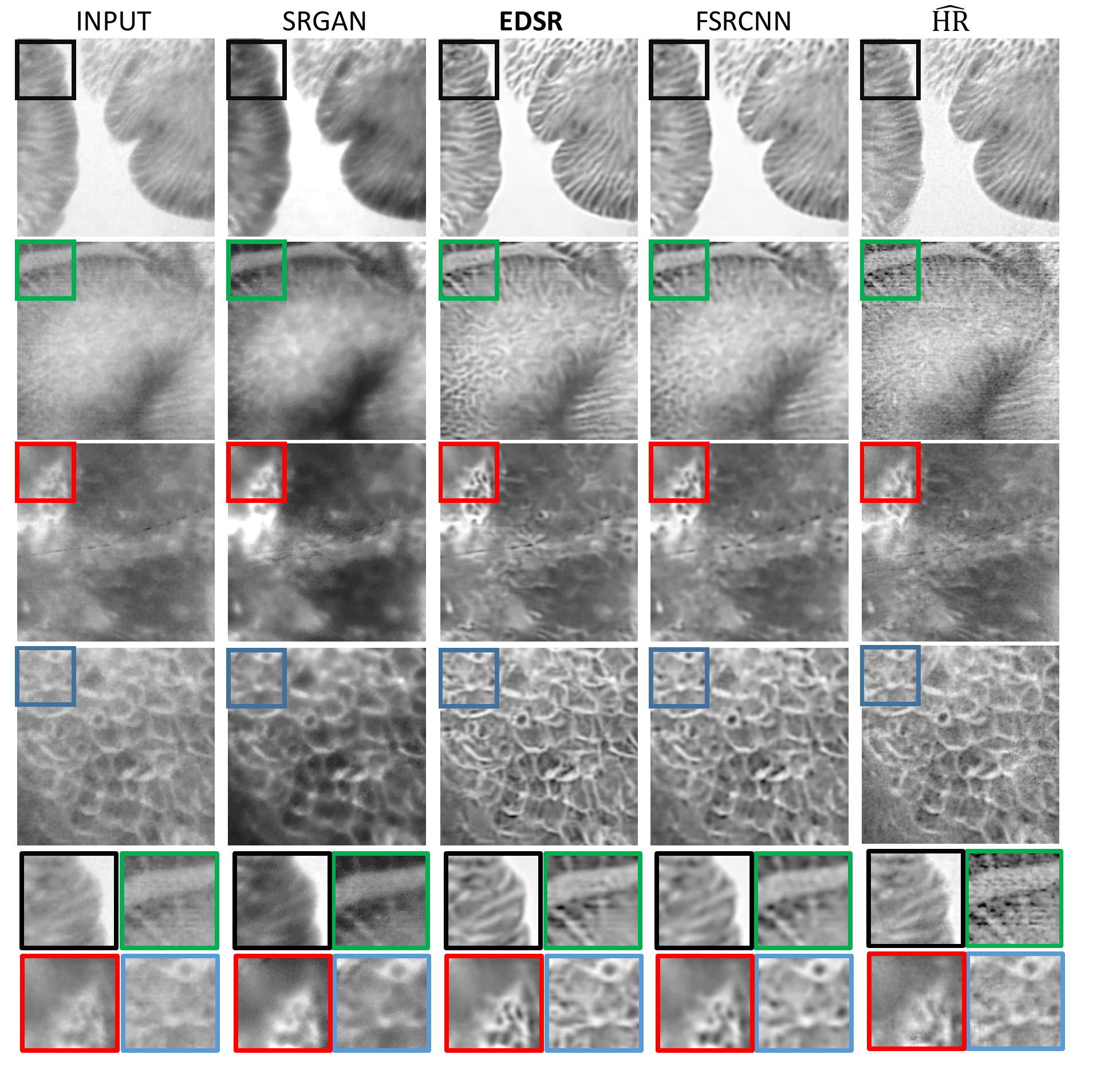}

    \caption{Example of visual results from the proposed approaches: Input (left), SRGAN (middle left), EDSR (middle) and FSRCNN (middle right) $\widehat{HR}$ (right)}
    \label{fig:SuperResolutionExamples}

\end{center}
\end{figure*}

\begin{table*}
\caption{Results of the proposed approach against state-of-the-art methods}
\label{table:StateOfTheArt}
\begin{center}
\resizebox{\textwidth}{!}{%

\begin{tabular}{c|>{\centering\arraybackslash}p{1.5cm}>{\centering\arraybackslash}p{1.7cm}>{\centering\arraybackslash}p{1.6cm}>{\centering\arraybackslash}p{1.6cm}>{\centering\arraybackslash}p{1.6cm}>{\centering\arraybackslash}p{1.7cm}}
	&Proposed&	Bayesian\cite{villena2013bayesian}&	PreTrained SRGAN	&	PreTrained EDSR L1& Wiener&	Contrast-enhancement	\\
	\hline
SSIM with $\widehat{HR}$&	0.8 $\pm$ 0.06&	\textbf{0.81 $\pm$ 0.06}&	0.79 $\pm$ 0.06& \textbf{0.81$\pm$ 0.06} &	0.77 $\pm$ 0.07&	0.65 $\pm$ 0.09	\\
$\Delta$ GCF with $\widehat{HR}$&	0.33 $\pm$ 0.31&	-0.26 $\pm$ 0.37&	-0.26 $\pm$ 0.36& -0.24$\pm$0.37&	-0.46 $\pm$ 0.48&	\textbf{0.81 $\pm$ 0.36}	\\
$\Delta$ GCF with LR&	0.57 $\pm$ 0.36&	-0.02 $\pm$ 0.01&	-0.01 $\pm$ 0.01	&0.00$\pm$0.01& -0.22 $\pm$ 0.24&	\textbf{1.06 $\pm$ 0.25}\\
\hline
$Tot_{cs}$&	\textbf{0.65}&	0.44&	0.40&0.44&	0.28&	0.50	\\

\end{tabular}
}
\end{center}
\end{table*}
To further validate our methodology, in Table~\ref{table:StateOfTheArt} the results obtained by the best model of our approach (EDSR trained on synthetic data with SSIM+L1 as loss function) were compared against other state-of-the-art SISR methodologies.
Specifically, in this experiment a Wiener deconvolution, a variational Bayesian inference approach with sparse and non-sparse priors~\cite{villena2013bayesian}, the SRGAN and EDSR networks pre-trained on natural images were considered. The Wiener deconvolution was assumed to have a Gaussian point-spread function with the parameter $\sigma$=2 estimated experimentally from the training set. Finally, the last column of Table~\ref{table:StateOfTheArt} includes the results of a contrast-enhancement approach obtained by sharpening the input with parameters similarly tuned on the trained set. Although our approach is not consistently outperforming the other on each individual quality score, when the combined score $Tot_{cs}$ is considered, our method outperforms the others by a large margin.

\subsection{Semi-quantitative analysis (MOS)}
\label{MOS}
To perform the MOS, nine independent experts were asked to evaluate 46 images each.
Full-size $LR_{org}$ were selected randomly from test set of $pCLE_{org}$, and used to generate SR reconstructions.
At each step, the SR images obtained by the three different methods (SRGAN, FSRCNN and EDSR) trained on synthetic data and a contrast-enhancement obtained by sharpening the input (used as a baseline) are shown to the user, in a randomly shuffled order. The input and the $\widehat{HR}$ are also displayed on the screen as references for the participants. For each of the four images the user assigns a score between 1 (strongly disagree) to 5 (strongly agree) on three different questions: 
\begin{itemize}
    \item \textit{Q1: Are there any artefacts/noise in the image?}
    \item \textit{Q2: Can you see an improvement in contrast with respect to the input?}
    \item \textit{Q3: Can you see an improvement in the details with respect to the input?} 
\end{itemize}

\begin{figure*}
\begin{center}

  \includegraphics[width=0.85\textwidth,trim={0 2.5cm 0 3.5cm},clip]{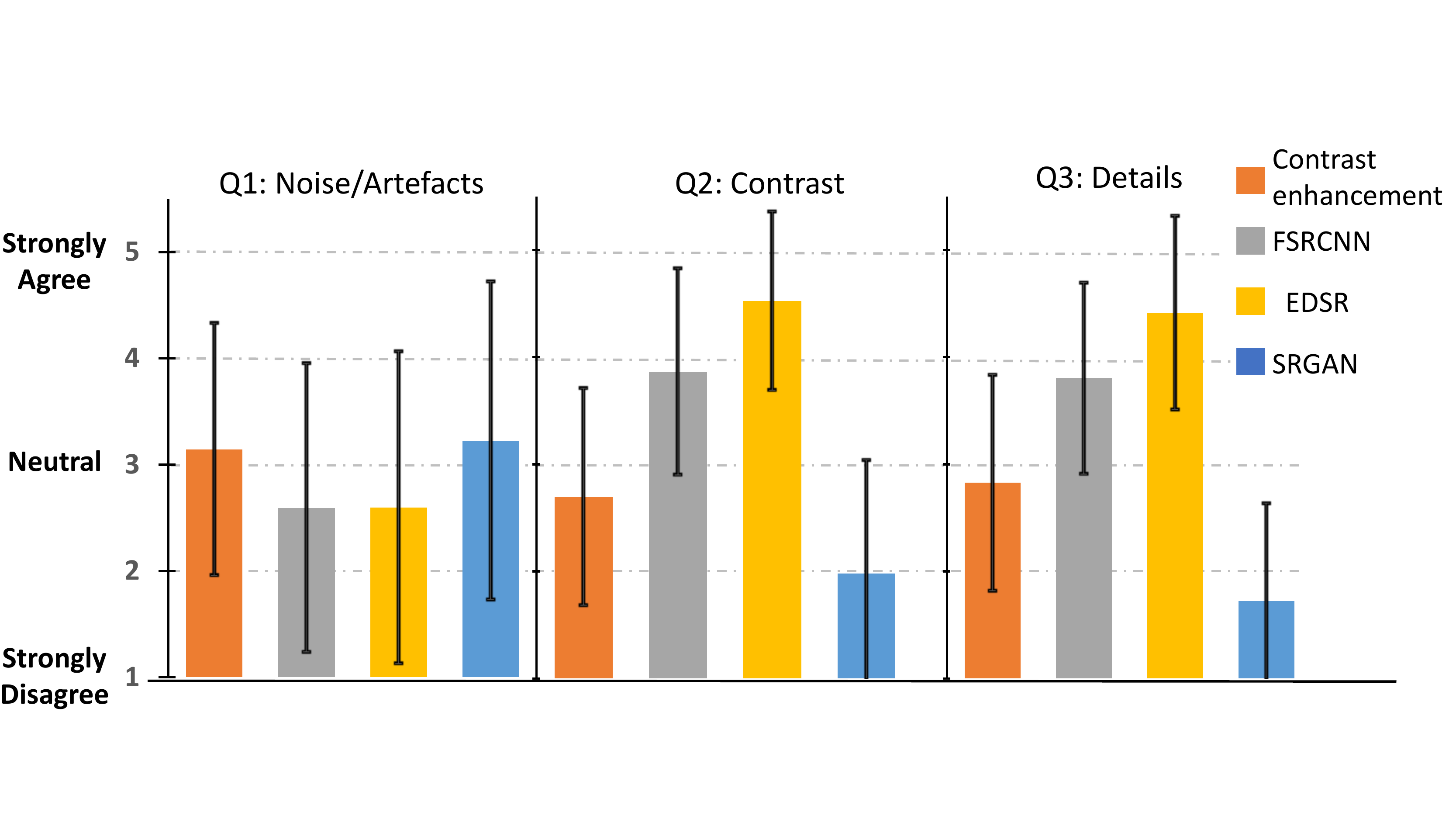}
\caption{Results of the MOS using a contrast-enhancement approach, FSRCNN, EDSR and SRGAN. The plots report the results on the 3 different questions.}
\label{fig:MOSResults}     
\end{center}
\end{figure*}

To make sure that the questions were correctly interpreted, each participant received a short training before starting the study. The results on the MOS are shown in Fig.~\ref{fig:MOSResults}. EDSR is the approach that achieves the best performance on Q2 and Q3. Instead based on Q1, both FRSCNN and EDSR do not introduce a significant amount of artefact or noise. The results of the MOS give us one more indication, which our training methodology allows improvements on the quality of the pCLE images. In Fig.~\ref{fig:SuperResolutionExamples} is shown a few examples of the obtained SR images using our proposed methodology.

%% file: disscusion.tex
\label{disscusion}
This work addresses the challenge of super-resolution for pCLE images. This is the first work to evaluate the potential of deep learning and exemplar-based super-resolution in pCLE context.
\par
The main contribution of this work is to overcome the challenge of lack of ground truth data. A novel methodology to produce pseudo-ground-truth exploiting an existing video-registration method, and simulating realistic LR image based on physical model of pCLE acquisition is proposed.
\par
The conclusions are that synthetic pCLE data can be used to train CNNs while applying them to real scenario data because of a physically-inspired simulation process that reduces the domain gap between real and simulated images.
\par
The robust IQA test based on the Structural Similarity (SSIM) and global contrast factor (GCF) score confirmed the improvement of obtained results in respects to the input image. An analysis of perceptual quality of images with a Mean Opinion Score (MOS) study recruiting nine independent pCLE experts showed that SR models give clinically interesting results. Experts perceived an improvement in the quality of the reconstructed images with respect to the input image without noting a significant increase in the amount of noise and artefacts. The quantitative and semi-quantitative user perception analysis provided consistent conclusions.\par

Providing a better quality of pCLE images might improve the decision process during the endoscopic examination. 
Further evaluation will focus on the temporal consistency of the super-resolution and will rely on histopathological confirmation to validate the authenticity of the generated details.